\documentclass[11pt,a4paper]{article}
\usepackage[hyperref]{naaclhlt2018}
\usepackage{times}
\usepackage{latexsym}
\usepackage{amsmath,amsfonts,amsthm,bm,amssymb}
\usepackage{graphicx}
\usepackage{url}
\usepackage{enumitem}
\usepackage{scrextend}
\usepackage{hhline}
\usepackage{url}

\aclfinalcopy 
\def\aclpaperid{543} 

\DeclareMathOperator{\EX}{\mathbb{E}}
\DeclareSymbolFontAlphabet{\mathcal}{symbols}
\DeclareSymbolFont{boldletters}{OT1}{cmr}{bx}{n}
\DeclareMathSymbol{x}{\mathalpha}{boldletters}{`x}
\DeclareMathSymbol{z}{\mathalpha}{boldletters}{`z}
\DeclareMathSymbol{c}{\mathalpha}{boldletters}{`c}
\DeclareMathSymbol{h}{\mathalpha}{boldletters}{`h}
\newcommand{\bmu}{\boldsymbol{\mathbf{\mu}}}
\newcommand{\bsigma}{\boldsymbol{\mathbf{\sigma}}}

\title{Rethinking Action Spaces for Reinforcement Learning in End-to-end Dialog Agents with Latent Variable Models}

\author{Tiancheng Zhao$^{1}$, Kaige Xie$^{2}$ and Maxine Eskenazi$^{1}$ \\
 $^{1}$Language Technologies Institute, Carnegie Mellon University \\
$^{2}$Shanghai Jiao Tong University \\
  {\tt \{tianchez, max+\}@cs.cmu.edu, lightyear0117@sjtu.edu.cn}}

\date{}

\begin{document}
\maketitle
\begin{abstract}
Defining action spaces for conversational agents and optimizing their decision-making process with reinforcement learning is an enduring challenge. Common practice has been to use handcrafted dialog acts, or the output vocabulary, e.g. in neural encoder decoders, as the action spaces. Both have their own limitations. This paper proposes a novel latent action framework that treats the action spaces of an end-to-end dialog agent as latent variables and develops unsupervised methods in order to induce its own action space from the data. Comprehensive experiments are conducted examining both continuous and discrete action types and two different optimization methods based on stochastic variational inference. Results show that the proposed latent actions achieve superior empirical performance improvement over previous word-level policy gradient methods on both DealOrNoDeal and MultiWoz dialogs. Our detailed analysis also provides insights about various latent variable approaches for policy learning and can serve as a foundation for developing better latent actions in future research.~\footnote{Data and code are available at \url{https://github.com/snakeztc/NeuralDialog-LaRL}}
\end{abstract}

\section{Introduction}
Optimizing dialog strategies in multi-turn dialog models is the cornerstone of building dialog systems that more efficiently solve real-world challenges, e.g. providing information~\cite{young2006using}, winning negotiations~\cite{lewis2017deal}, improving engagement~\cite{li2016deep} etc. A classic solution employs reinforcement learning (RL) to learn a dialog policy that models the optimal action distribution conditioned on the dialog state~\cite{williams2007partially}. However, since there are infinite human language possibilities, an enduring challenge has been to define what the action space is. For traditional modular systems, the action space is defined by hand-crafted semantic representations such as dialog acts and slot-values~\cite{raux2005let,chen2013unsupervised} and the goal is to obtain a dialog policy that chooses the best hand-crafted action at each dialog turn. But it is limited because it can only handle simple domains whose entire action space can be captured by hand-crafted representations~\cite{walker2000application,su2017sample}. This cripples a system's ability to handle conversations in complex domains.

Conversely, end-to-end (E2E) dialog systems have removed this limit by directly learning a response generation model conditioned on the dialog context using neural networks~\cite{vinyals2015neural,sordoni2015neural}. To apply RL to E2E systems, the action space is typically defined as the entire vocabulary; every response output word is considered to be an action selection step~\cite{li2016deep}, which we denote as the word-level RL. Word-level RL, however, has been shown to have several major limitations in learning dialog strategies. The foremost one is that direct application of word-level RL leads to degenerate behavior: the response decoder deviates from human language and generates utterances that are incomprehensible~\cite{lewis2017deal,das2017learning,kottur2017natural}. A second issue is that since a multi-turn dialog can easily span hundreds of words, word-level RL suffers from credit assignment over a long horizon, leading to slow and sub-optimal convergence~\cite{kaelbling1996reinforcement,he2018decoupling}.   

This paper proposes Latent Action Reinforcement Learning (LaRL), a novel framework that overcomes the limitations of word-level RL for E2E dialog models, marrying the benefits of a traditional modular approach in an unsupervised manner. The key idea is to develop E2E models that can invent their own discourse-level actions. These actions must be expressive enough to capture response semantics in complex domains (i.e. have the capacity to represent a large number of actions), thus decoupling the discourse-level decision-making process from natural language generation. Then any RL technique can be applied to this induced action space in the place of word-level output. We propose a flexible latent variable dialog framework and investigate several approaches to inducing latent action space from natural conversational data. We further propose (1) a novel training objective that outperforms the typical evidence lower bound used in dialog generation and (2) an attention mechanism for integrating discrete latent variables in the decoder to better model long responses.

We test this on two datasets,  DealOrNoDeal~\cite{lewis2017deal} and MultiWoz~\cite{budzianowski2018multiwoz}, to answer two key questions: (1) what are the advantages of LaRL over Word-level RL and (2) what effective methods can induce this latent action space. Results show that LaRL is significantly more effective than word-level RL for learning dialog policies and it does not lead to incomprehensible language generation. Our models achieve 18.2\% absolute improvement over the previous state-of-the-art on MultiWoz and discover novel and diverse negotiation strategies on DealOrNoDeal. Besides strong empirical improvement, our model analysis reveals novel insights, e.g. it is crucial to reduce the exposure bias in the latent action space and discrete latent actions are more suitable than continuous ones to serve as action spaces for RL dialog agents.

\section{Related Work}
Prior RL research in modular dialog management has focused on policy optimization over hand-crafted action spaces in task-oriented domains~\cite{walker2000application,young2007hidden}. A dialog manager is formulated as a Partially Observable Markov Decision Process (POMDP)~\cite{young2013pomdp}, where the dialog state is estimated via dialog state tracking models from the raw dialog context~\cite{lee2013structured,henderson2014word,ren2018towards}. RL techniques are then used to find the optimal dialog policy~\cite{gasic2014gaussian,su2017sample,williams2017hybrid}. Recent deep-learning modular dialog models have also explored joint optimization over dialog policy and state tracking to achieve stronger performance~\cite{wen2016network,zhao2016towards,liu2017end}.

A related line of work is reinforcement learning for E2E dialog systems. Due to the flexibility of encoder-decoder dialog models, prior work has applied reinforcement learning to more complex domains and achieved higher dialog-level rewards, such as open-domain chatting~\cite{li2016deep,serban2017deep}, negotiation~\cite{lewis2017deal}, visual dialogs~\cite{das2017learning}, grounded dialog~\cite{mordatch2017emergence} etc. As discussed in Section 1, these methods consider the output vocabulary at every decoding step to be the action space; they suffer from limitations such as deviation from natural language and sub-optimal convergence. 

Finally, research in latent variable dialog models is closely related to our work, which strives to learn meaningful latent variables for E2E dialog systems. Prior work has shown that learning with latent variables leads to benefits like diverse response decoding~\cite{serban2017hierarchical,zhao2017learning,cao2017latent}, interpretable decision-making~\cite{wen2017latent,zhao2018unsupervised} and zero-shot domain transfer~\cite{zhao2018zero}. Also, driven by similar motivations of this work, prior studies have explored to utilize a coarse discrete node, either handcrafted or learned, to decouple the word generation process from dialog policy in E2E systems for better dialog policy~\cite{he2018decoupling,yarats2017hierarchical}. Our work differs from prior work for two reasons: (1) latent action in previous work is only auxiliary, small-scale and mostly learned in a supervised or semi-supervised setting. This paper focuses on unsupervised learning of latent variables and learns variables that are expressive enough to capture the entire action space by itself. (2) to our best knowledge, our work is the first comprehensive study of the use of latent variables for RL policy optimization in dialog systems.

\section{Baseline Approach}
\label{sec:baseline}
\begin{figure*}[ht]
    \centering
    \includegraphics[width=16cm]{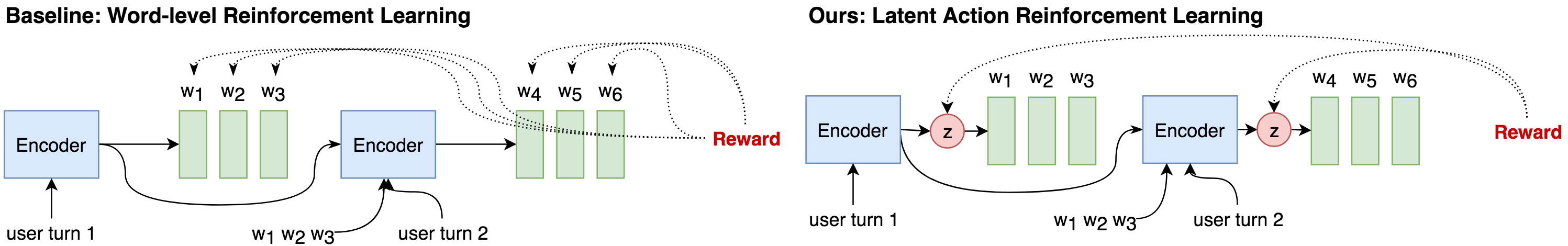}
    \caption{High-level comparison between word-level and latent-action reinforcement learning in a sample multi-turn dialog. The green boxes denote the decoder network used to generate the response given the latent code $z$. Dashed line denotes places where policy gradients from task rewards are applied to the model. }
    \label{fig:overview}
\end{figure*}

E2E response generation can be treated as a conditional language generation task, which uses neural encoder-decoders~\cite{cho2014learning} to model the conditional distribution $p(x|c)$ where $c$ is the observed dialog context and $x$ is the system's response to the context. The format of the dialog context is domain dependent. It can vary from textual raw dialog history~\cite{vinyals2015neural} to visual and textual context~\cite{das2017learning}. Training with RL usually has 2 steps: supervised pre-training and policy gradient reinforcement learning~\cite{williams2016end,dhingra2017towards,li2016deep}. Specifically, the supervised learning step maximizes the log likelihood on the training dialogs, where $\theta$ is the model parameter:
\begin{equation}
    \mathcal{L}_{SL}({\theta}) = \EX_{x, c}[\log p_\theta(x|c)]
\end{equation}
Then the following RL step uses policy gradients, e.g. the REINFORCE algorithm~\cite{williams1992simple} to update the model parameters with respect to task-dependent goals. We assume that we have an environment that the dialog agent can interact with and that there is a turn-level reward $r_t$ at every turn $t$ of the dialog. We can then write the expected discounted return under a dialog model $\theta$ as $J(\theta) = \EX[\sum_0^T \gamma^t r_t]$, where $\gamma  \in [0, 1]$ is the discounting factor and $T$ is the length of the dialog. Often a baseline function $b$ is used to reduce the variance of the policy gradient~\cite{greensmith2004variance}, leading to $R_t = \sum_{k=0} ^ {T-t} \gamma^k (r_{t+k} - b)$. 

\textbf{Word-level Reinforcement Learning}: as shown in Figure~\ref{fig:overview}, the baseline approach treats every output word as an action step and its policy gradient is:
\begin{equation}
    \nabla_\theta J(\theta) = \EX_\theta [\sum_{t=0}^T \sum_{j=0}^{U_t} R_{tj}  \nabla_\theta \log p_\theta (w_{tj}|w_{< tj}, c_t)]
\label{eq:wordRL}
\end{equation}
where $U_t$ is the number of tokens in the response at turn $t$ and $j$ is the word index in the response. It is evident that Eq~\ref{eq:wordRL} has a very large action space, i.e. $|V|$ and a long learning horizon, i.e. $ TU$. Prior work has found that the direct application of Eq~\ref{eq:wordRL} leads to divergence of the decoder. The common solution is to alternate with supervised learning with Eq~\ref{eq:wordRL} at a certain ratio~\cite{lewis2017deal}. We denote this ratio as RL:SL=A:B, which means for every A policy gradient updates, we run B supervised learning updates. We use RL:SL=off for the case where only policy gradients are used and no supervised learning is involved.

\section{Latent Action Reinforcement Learning}
\label{sec:method}
We now describe the proposed LaRL framework. As shown in Figure~\ref{fig:overview}, a latent variable $z$ is introduced in the response generation process. The conditional distribution is factorized into $p(x|c) = p(x|z)p(z|c)$ and the generative story is: (1) given a dialog context $c$ we first sample a latent action $z$ from $p_{\theta_e}(z|c)$ and (2) generate the response by sampling $x$ based on $z$ via $p_{\theta_d}(x|z)$, where $p_{\theta_e}$ is the dialog encoder network and $p_{\theta_d}$ is the response decoder network. Given the above setup, LaRL treats the latent variable $z$ as its action space instead of outputting words in response $x$. We can now apply REINFORCE in the latent action space:
\begin{equation}
    \nabla_\theta J(\theta) = \EX_\theta [\sum_{t=0}^T R_t \log p_{\theta}(z|c_t) ]
\label{eq:laRL}
\end{equation}
Compared to Eq~\ref{eq:wordRL}, LaRL differs by: 
\begin{itemize}
    \item Shortens the horizon from $TU$ to $T$.
    \item Latent action space is designed to be low-dimensional, much smaller than $V$.
    \item The policy gradient only updates the encoder $\theta_e$ and the decoder $\theta_d$ stays intact.
\end{itemize}
These properties reduce the difficulties for dialog policy optimization and decouple high-level decision-making from natural language generation. The $p_{\theta_e}$ are responsible for choosing the best latent action given a context $c$ while $p_{\theta_d}$ is only responsible for transforming $z$ into the surface-form words. Our formulation also provides a flexible framework for experimenting with various types of model learning methods. In this paper, we focus on two key aspects: the type of latent variable $z$ and optimization methods for learning $z$ in the supervised pre-training step.

\subsection{Types of Latent Actions}
Two types of latent variables have been used in previous research: continuous isotropic Gaussian distribution~\cite{serban2017hierarchical} and multivariate categorical distribution~\cite{zhao2018unsupervised}. These two types are both compatible with our LaRL framework and can be defined as follows:

\textbf{Gaussian Latent Actions} follow $M$ dimensional multivariate Gaussian distribution with a diagonal covariance matrix, i.e. $z \sim \mathcal{N}(\bmu , \bsigma^2\textbf{I})$. Let the encoder $p_{\theta_e}$ consist of two parts: a context encoder $\mathcal{F}$, a neural network that encodes the dialog context $c$ into a vector representation $h$, and a feed forward network $\pi$ that projects $h$ into $\bmu$ and $\bsigma$. The process is defined as follows:
\begin{align}
    &h = \mathcal{F}(c) \\
    &\begin{bmatrix}
        \bmu \\
        \log(\bsigma^{2})
    \end{bmatrix} = \pi(h) \\ 
    &p(x|z) = p_{\theta_d}(z) \quad z \sim \mathcal{N}(\bmu, \bsigma^2 \textbf{I})
\end{align}
where the sampled $z$ is used as the initial state of the decoder for response generation. Also we use $p_{\theta}(z|c) = \mathcal{N}(z ; \bmu, \bsigma^2 \textbf{I})$ to compute the policy gradient update in Eq~\ref{eq:laRL}.

\textbf{Categorical Latent Actions} are $M$ independent K-way categorical random variables. Each $z_m$ has its own token embeddings to map latent symbols into vector space $\bm{E}_m \in \mathbb{R}^{K \times D}$ where $m \in [1, M]$ and $D$ is the embedding size. Thus $M$ latent actions can represent exponentially, $K^M$, unique combinations, making it expressive enough to model dialog acts in complex domains. Similar to Gaussian Latent Actions, we have
\begin{align}
    &h = \mathcal{F}(c) \\
    &p(Z_m|c) = \text{softmax}(\pi_m(h))  \\
    &p(x|z) = p_{\theta_d}(\bm{E}_{1:M}(z_{1:M})) \quad z_m \sim p(Z_m|c) 
\end{align}
For the computing policy gradient in Eq~\ref{eq:laRL}, we have $p_{\theta}(z|c) = \prod_{m=1}^{M} p(Z_m=z_m|c)$

Unlike Gaussian latent actions, a matrix $\mathbb{R}^{M \times D}$ comes after the embedding layers $\bm{E}_{1:M}(z_{1:M})$, whereas the decoder's initial state is a vector of size $\mathbb{R}^{D}$. Previous work integrated this matrix with the decoder by summing over the latent embeddings, i.e.  $x = p_{\theta_d}(\sum_1^M \bm{E}_m(z_m))$, denoted as \textbf{Summation Fusion} for later discussion~\cite{zhao2018unsupervised}. A limitation of this method is that it could lose fine-grained order information in each latent dimension and have issues with long responses that involve multiple dialog acts. Therefore, we propose a novel method, \textbf{Attention Fusion}, to combine categorical latent actions with the decoder. We apply the attention mechanism~\cite{luong2015effective} over latent actions as the following. Let $i$ be the step index during decoding. Then we have:
\begin{align}
    \alpha_{mi} &= \text{softmax}(h_i^{T} \bm{W}_a \bm{E}_m(z_m)) \\
    c_i &= \sum_{m=1}^M \alpha_{mi} \bm{E}_m(z_m) \\
    \widetilde{h_i} &= \text{tanh}(\bm{W}_s 
    \begin{bmatrix}
        h_i \\
        c_i
    \end{bmatrix}) \\
    p(w_i|h_i, c_i) &= \text{softmax}(W_o \widetilde{h_i})
\end{align}
The decoder's next state is updated by $h_{i+1} = \text{RNN}(h_i,w_{i+1}), \widetilde{h_i})$ and $h_0$ is computed via summation-fusion. Thus attention fusion lets the decoder focus on different latent dimensions at each generation step. 

\subsection{Optimization Approaches}
\textbf{Full ELBO}: Now given a training dataset $\{x, c\}$, our base optimization method is via stochastic variational inference by maximizing the evidence lowerbound (ELBO), a lowerbound on the data log likelihood:
\begin{equation}
    \mathcal{L}_{full}(\theta) = p_{q(z|x, c)}(x|z) - \text{D}_{\text{KL}}[q(z|x, c) \| p(z|c)]    
    \label{eq:elbo}
\end{equation}
where $q_\gamma(z|x, c)$ is a neural network that is trained to approximate the posterior distribution $q(z|x, c)$ and $p(z|c)$ and $p(x|z)$ are achieved by $\mathcal{F}$, $\pi$ and $p_{\theta_d}$. For Gaussian latent actions, we use the reparametrization trick~\cite{kingma2013auto} to backpropagate through Gaussian latent actions and the Gumbel-Softmax~\cite{jang2016categorical} to backpropagate through categorical latent actions.

\textbf{Lite ELBO}: a major limitation is that Full ELBO can suffer from exposure bias at latent space, i.e. the decoder only sees $z$ sampled from  $q(z|x, c)$ and never experiences $z$ sampled from $p_\theta(z|c)$, which is always used at testing time. Therefore, in this paper, we propose a simplified ELBO for encoder-decoder models with stochastic latent variables:
\begin{equation}
    \mathcal{L}_{lite}(\theta)  = p_{p(z|c)}(x|z) - \beta \text{D}_{\text{KL}}[p(z|c)) \| p(z)]
\label{eq:simple_elbo}
\end{equation}
Essentially this simplified objective sets the posterior network the same as our encoder, i.e. $q_\gamma(z|x,c)=p_{\theta_e}(z|c)$, which makes the KL term in Eq~\ref{eq:elbo} zero and removes the issue of exposure bias. But this leaves the latent spaces unregularized and our experiments show that if we only maximize $p_{p(z|c)}(x|z)$ there is overfitting. For this, we add the additional regularization term $\beta \text{D}_{\text{KL}}[p(z|c)) \| p(z)]$ that encourages the posterior be similar to certain prior distributions and $\beta$ is a hyper-parameter between 0 and 1. We set the $p(z)$ for categorical latent actions to be uniform, i.e. $p(z)=1/K$, and set the prior for Gaussian latent actions to be $\mathcal{N}(\bm{0}, \textbf{I})$, which we will show that are effective.

\section{Experiment Settings}
\label{sec:setup}
\subsection{DealOrNoDeal Corpus and RL Setup}

DealOrNoDeal is a negotiation dataset that contains 5805 dialogs based on 2236 unique scenarios~\cite{lewis2017deal}. We hold out 252 scenarios for testing environment and randomly sample 400 scenarios from the training set for validation. The results are evaluated from 4 perspectives: Perplexity (PPL), Reward, Agree and Diversity. PPL helps us to identify which model produces the most human-like responses, while Reward and Agree evaluate the model's negotiation strength. Diversity indicates whether the model discovers a novel discourse-level strategy or just repeats dull responses to compromise with the opponent. We closely follow the original paper and use the same reward function and baseline calculation. At last, to have a fair comparison, all the compared models shared the identical judge model and user simulator, which are a standard hierarchical encoder-decoder model trained with Maximum Likelihood Estimation (MLE).

\subsection{Multi-Woz Corpus and Novel RL Setup}
Multi-Woz is a slot-filling dataset that contains 10438 dialogs on 6 different domains. 8438 dialogs are for training and 1000 each are for validation and testing. Since no prior user simulator exists for this dataset, for a fair comparison with the previous state-of-the-art we focus on the Dialog-Context-to-Text Generation task proposed in~\cite{budzianowski2018multiwoz}. This task assumes that the model has access to the ground-truth dialog belief state and is asked to generate the next response at every system turn in a dialog. The results are evaluated from 3 perspectives: BLEU, Inform Rate and Success Rate. The BLEU score checks the response-level lexical similarity, while Inform and Success Rate measure whether the model gives recommendations and provides all the requested information at dialog-level. Current state-of-the-art results struggle in this task and MLE models only achieve 60\% success~\cite{budzianowski2018multiwoz}. To transform this task into an RL task, we propose a novel extension to the original task as follows:
\begin{enumerate}
    \item For each RL episode, randomly sample a dialog from the training set
    \item Run the model on every system turn, and do not alter the original dialog context at every turn given the generated responses.
    \item Compute Success Rate based on the generated responses in this dialog.
    \item Compute policy gradient using Eq~\ref{eq:laRL} and update the parameters.
\end{enumerate}
This setup creates a variant RL problem that is similar to the Contextual Bandits~\cite{langford2008epoch}, where the goal is to adjust its parameters to generate responses that yield better Success Rate. Our results show that this problem is challenging and that word-level RL falls short.

\subsection{Language Constrained Reward (LCR) curve for Evaluation}
It is challenging to quantify the performance of RL-based neural generation systems because it is possible for a model to achieve high task reward and yet not generate human language~\cite{das2017learning}. Therefore, we propose a novel measure, the Language Constrained Reward (LCR) curve as an additional robust measure. The basic idea is to use an ROC-style curve to visualize the tradeoff between achieving higher reward and being faithful to human language. Specifically, at each checkpoint $i$ over the course of RL training, we record two measures: (1) the PPL of a given model on the test data $p_i=\text{PPL}(\theta_i)$ and (2) this model's average cumulative task reward in the test environment $R^t_i$. After RL training is complete, we create a 2D plot where the x-axis is the maximum PPL allowed, and the y-axis is the best achievable reward within the PPL budget in the testing environments:
\begin{equation}
    y = \text{max}_i R_i^t \quad \text{subject to}\quad p_i < \it{\mathnormal{x}}
\end{equation}

As a result, a perfect model should lie in the upper left corner whereas a model that sacrifices language quality for higher reward will lie in the lower right corner. Our results will show that the LCR curve is an informative and robust measure for model comparison. 

\section{Results: Latent Actions or Words?}
\label{sec:results}
We have created 6 different variations of latent action dialog models under our LaRL framework.
\begin{table}[ht]
\centering
\small
\begin{tabular}{p{0.24\linewidth}|p{0.22\linewidth}p{0.1\linewidth}p{0.2\linewidth}}\hline
Model            & Var Type         & Loss                  & Integration \\ \hline
Gauss            & Gaussian         & $\mathcal{L}_{full}$  & /  \\
Cat              & Categorical      & $\mathcal{L}_{full}$  & sum  \\
AttnCat          & Categorical      & $\mathcal{L}_{full}$  & attn \\
LiteGauss      & Gaussian         & $\mathcal{L}_{lite}$    & /  \\
LiteCat        & Categorical      & $\mathcal{L}_{lite}$    & sum  \\
LiteAttnCat    & Categorical      & $\mathcal{L}_{lite}$    & attn \\ \hline
\end{tabular}
\caption{All proposed variations of LaRL models.}
\label{tbl:models}
\end{table}
To demonstrate the advantages of LaRL, during the RL training step, we set RL:SL=off for all latent action models, while the baseline word-level RL models are free to tune RL:SL for best performance. For latent variable models, their perplexity is estimated via Monte Carlo $p(x|c) \approx \EX_{p(z|c)}[p(x|z)p(z|c)]$. For the sake of clarity, this section only compares the best performing latent action models to the best performing word-level models and focuses on the differences between them. A detailed comparison of the 6 latent space configurations is addressed in Section~\ref{sec:analysis}.

\subsection{DealOrNoDeal}
The baseline system is a hierarchical recurrent encoder-decoder (HRED) model~\cite{serban2016building} that is tuned to reproduce results from~\cite{lewis2017deal}. Word-level RL is then used to fine-tune the pre-trained model with RL:SL=4:1. On the other hand, the best performing latent action model is LiteCat. Best models are chosen based on performance on the validation environment.
\begin{table}[ht]
\begin{tabular}{p{0.09\textwidth}|p{0.05\textwidth}p{0.07\textwidth}p{0.07\textwidth}p{0.09\textwidth}} \hline
                 & PPL             & Reward  & Agree\%       & Diversity \\  \hline
Baseline         & 5.23   & 3.75    & 59            & 109      \\
LiteCat          & 5.35            & 2.65    & 41            & 58 \\ \hline
Baseline +RL      & 8.23            & \textbf{7.61}     & 86           & 5  \\ 
LiteCat +RL       & \textbf{6.14}            & 7.27    & \textbf{87}  & \textbf{202}      \\ \hline 
\end{tabular}
\caption{Results on DealOrNoDeal. Diversity is measured by the number of unique responses the model used in all scenarios from the test data.}
\label{tbl:deal}
\end{table}

\begin{figure}[ht]
    \centering
    \includegraphics[width=0.45\textwidth]{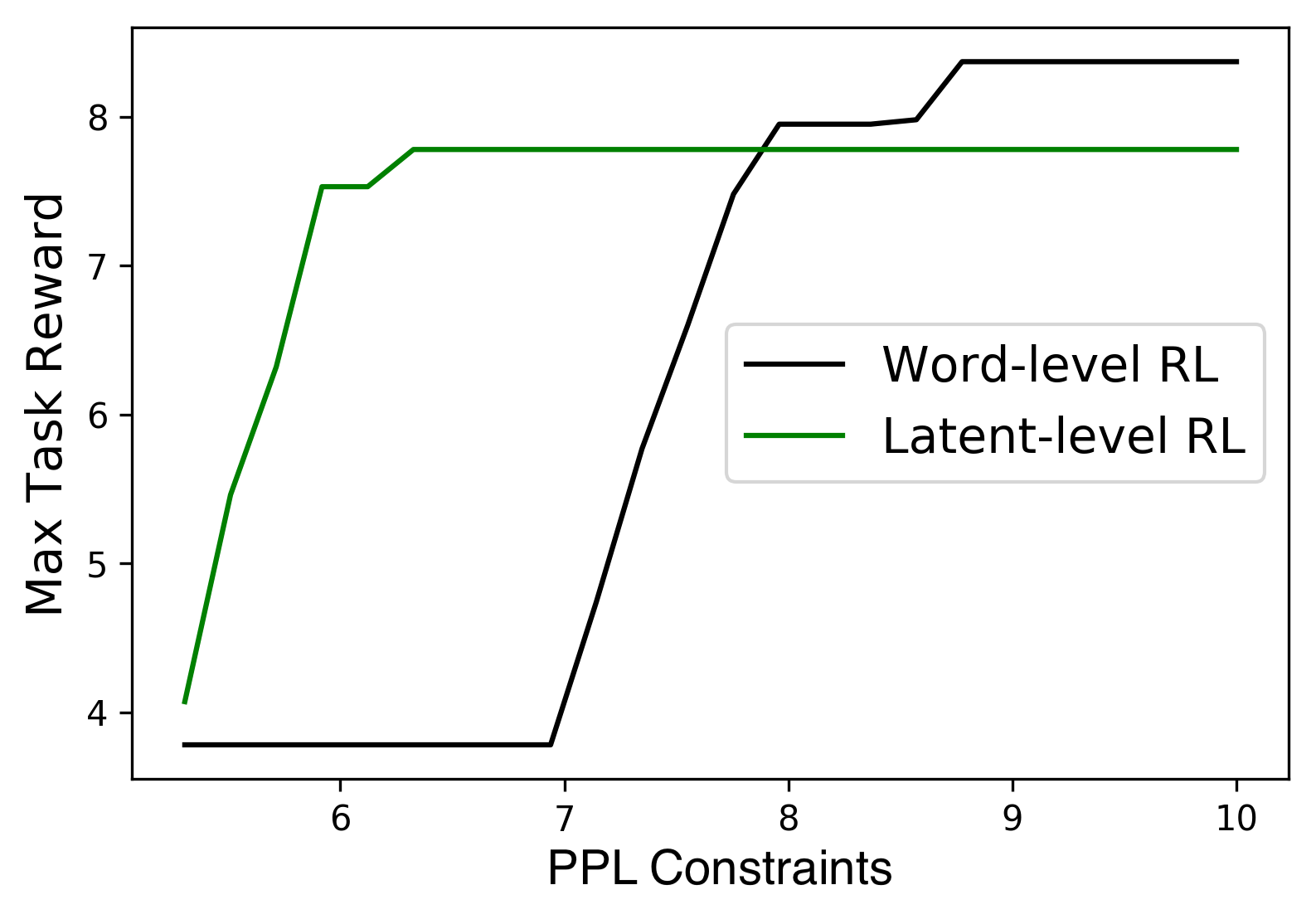}
    \caption{LCR curves on DealOrNoDeal dataset.}
    \label{fig:deal_lcr}
\end{figure}

The results are summarized in Table~\ref{tbl:deal} and Figure~\ref{fig:deal_lcr} shows the LCR curves for the baseline with the two best models plus LiteAttnCat and baseline without RL:SL. From Table~\ref{tbl:deal}, it appears that the word-level RL baseline performs better than LiteCat in terms of rewards. However, Figure~\ref{fig:deal_lcr} shows that the two LaRL models achieve strong task rewards with a much smaller performance drop in language quality (PPL), whereas the word-level model can only increase its task rewards by deviating significantly from natural language.  

\begin{figure}[ht]
    \centering
    \includegraphics[width=0.3\textwidth]{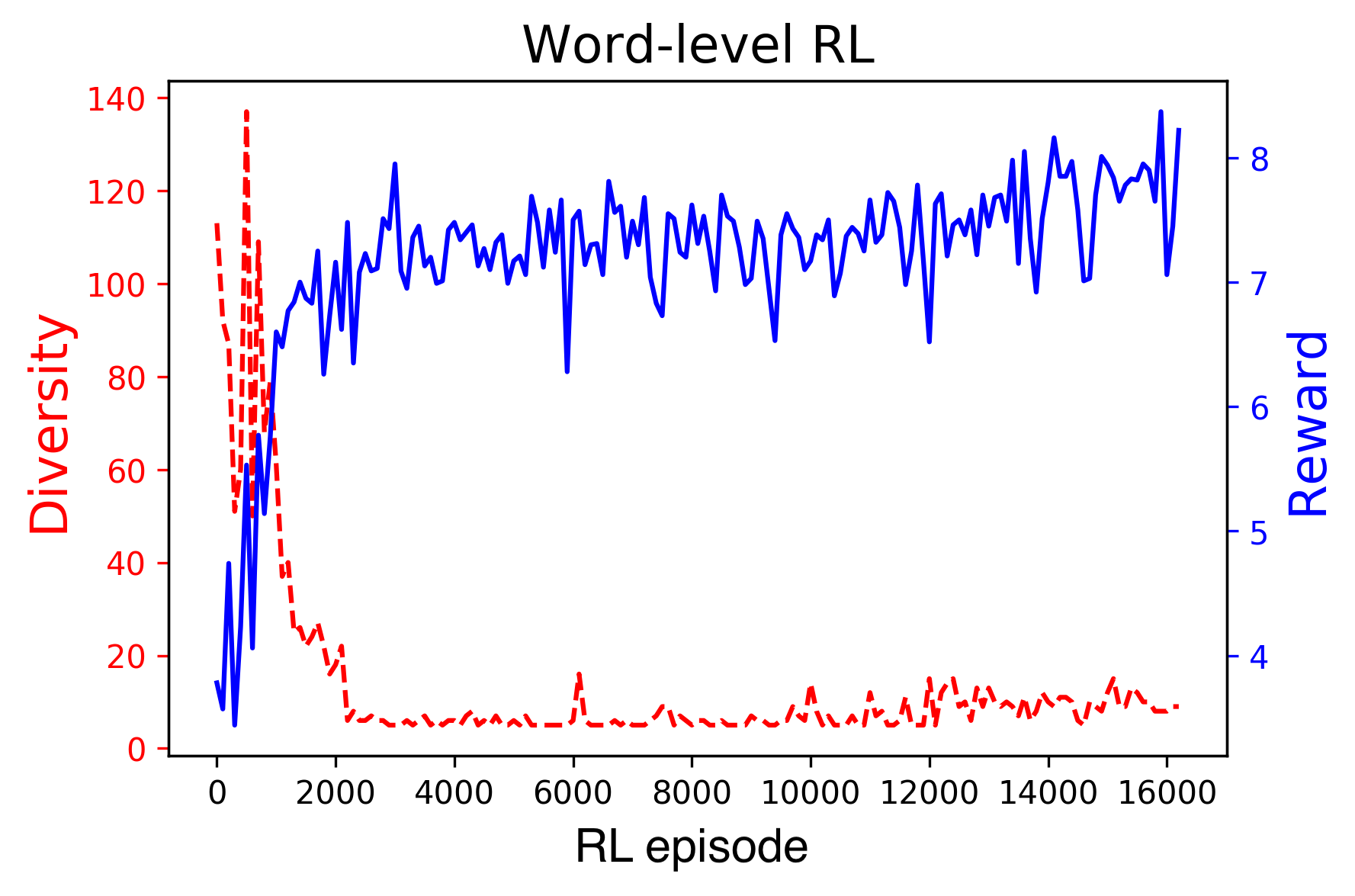}
    \includegraphics[width=0.3\textwidth]{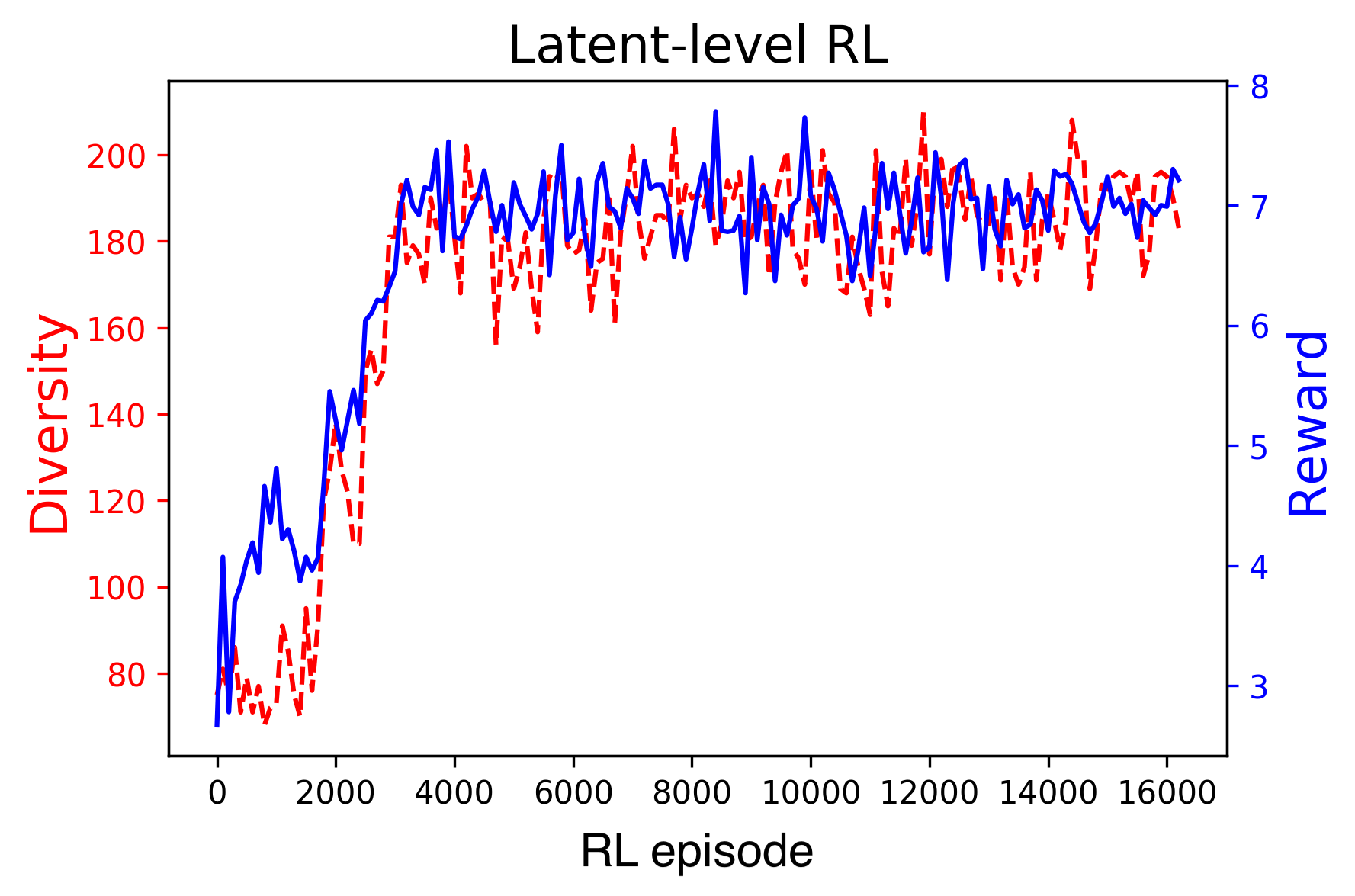}
    \caption{Response diversity and task reward learning curve over the course of RL training for both word RL:SL=4:1 (left) and LiteCat (right).}
    \label{fig:deal_diversity}
\end{figure}

Closer analysis shows the word-level baseline severely overfits to the user simulator. The caveat is that the word-level models have in fact discovered a loophole in the simulator by insisting on 'hat' and 'ball' several times and the user model eventually yields to agree to the deal. This is reflected in the diversity measure, which is the number of unique responses that a model uses in all 200 testing scenarios. As shown in Figure~\ref{fig:deal_diversity}, after RL training, the diversity of the baseline model drops to only 5. It is surprising that the agent can achieve high reward with a well-trained HRED user simulator using only 5 unique utterances. On the contrary, LiteCat increases its response diversity after RL training from 58 to 202, suggesting that LiteCat discovers novel discourse-level strategies in order to win the negotiation instead of exploiting local loopholes in the same user simulator. Our qualitative analysis confirms this when we observe that our LiteCat model is able to use multiple strategies in negotiation, e.g. elicit preference question, request different offers, insist on key objects etc. See supplementary material for example conversations.  

\subsection{MultiWoz}
For MultiWoz, we reproduce results from~\cite{budzianowski2018multiwoz} as the baseline. After RL training, the best LaRL model is LiteAttnCat and the best word-level model is word RL:SL=off.
\begin{table}[ht]
\begin{tabular}{p{0.11\textwidth}|p{0.05\textwidth}p{0.05\textwidth}p{0.07\textwidth}p{0.07\textwidth}} \hline
                 & PPL    & BLEU  & Inform   &  Success \\  \hline
Human            & /       & /    & 90\%        & 82.3\%     \\ \hline
Baseline         & 3.98   & 18.9   & 71.33\%        & 60.96\%       \\
LiteAttnCat      & 4.05    & 19.1   & 67.98\%     & 57.36\%    \\\hline
Baseline +RL     & 17.11   & 1.4   & 80.5\%      & 79.07\%    \\ 
LiteAttnCat +RL  & \textbf{5.22}    & \textbf{12.8}    & \textbf{82.78\%}      & \textbf{79.2\%}   \\ \hline 
\end{tabular}
\caption{Main results on MultiWoz test set. RL models are chosen based on performance on the validation set.}
\label{tbl:multi_woz_main}
\end{table}
Table~\ref{tbl:multi_woz_main} shows that LiteAttnCat is on par with the baseline in the supervised learning step, showing that multivariate categorical latent variables alone are powerful enough to match with continuous hidden representations for modeling dialog actions. For performance after RL training, LiteAttnCat achieves near-human performance in terms of success rate and inform rate, obtaining 18.24\% absolute improvement over the MLE-based state-of-the-art~\cite{budzianowski2018multiwoz}. More importantly, perplexity only slightly increases from 4.05 to 5.22. On the other hand, the word-level RL's success rate also improves to 79\%, but the generated responses completely deviate from natural language, increasing perplexity from 3.98 to 17.11 and dropping BLEU from 18.9 to 1.4. 
\begin{figure}[ht]
    \centering
    \includegraphics[width=0.49\textwidth]{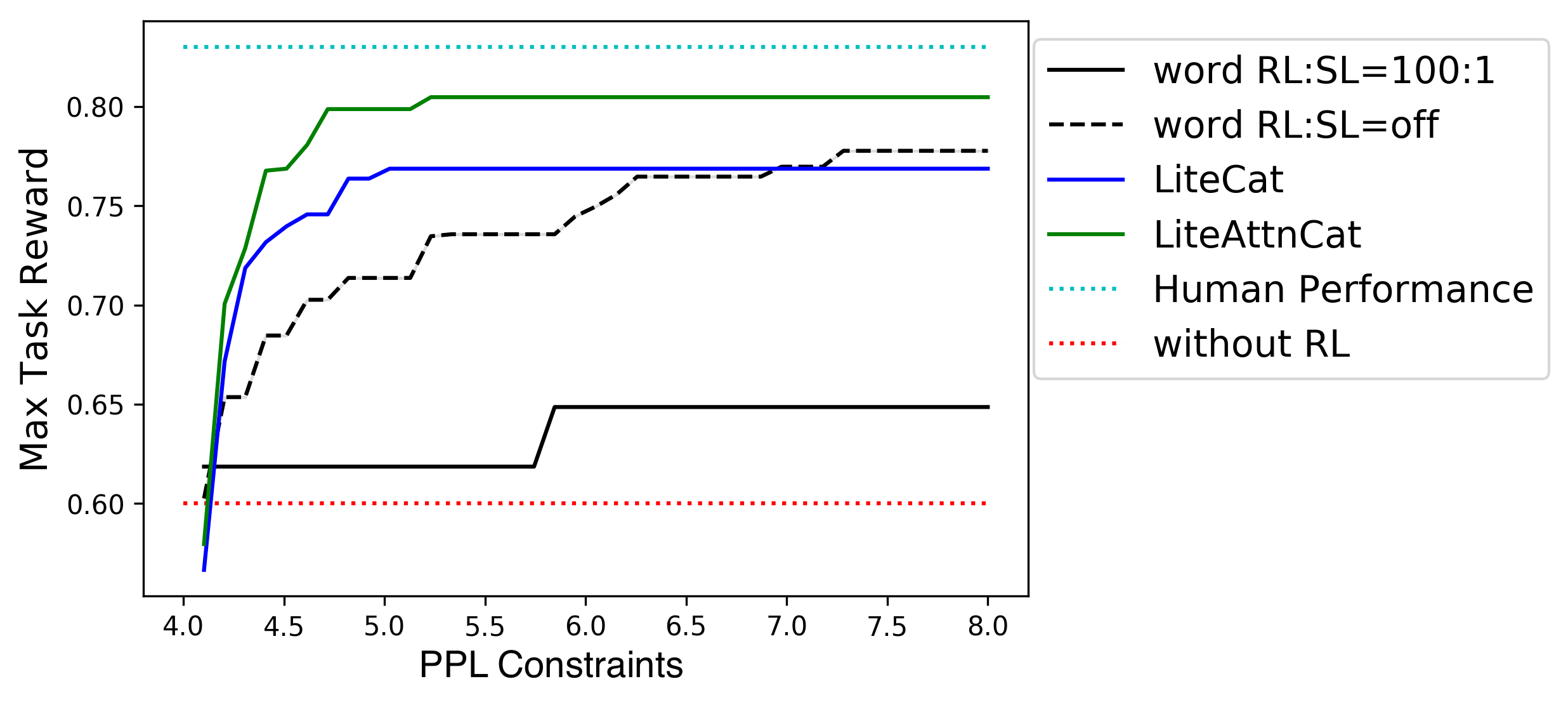}
    \caption{LCR curves on the MultiWoz dataset.}
    \label{fig:woz_lcr}
\end{figure}

Figure~\ref{fig:woz_lcr} shows the LCR curves for MultiWoz, with a trend similar to the previous section: the word-level models can only achieve task reward improvement by sacrificing their response decoder PPL. Figure~\ref{fig:woz_lcr} also shows the LCR curve for the baseline trained with RL:SL=100:1, hoping that supervised learning can force the model to conform to natural language. While PPL and BLEU are indeed improved, it also limits final reward performance. The latent-level models, on the contrary, do not suffer from this tradeoff. We also observe that LiteAttnCat consistently outperforms LiteCat on MultiWoz, confirming the effectiveness of Attention Fusion for handling long dialog responses with multiple entities and dialog acts. Lastly, Table~\ref{tbl:multiwoz_example} qualitatively exhibits the generation differences between the two approaches. The RL:SL=off model learns to continuously output entities to fool the evaluation script for high success rate, whereas LiteCatAttn learns to give more information while maintaining the language quality.

\begin{table}[ht]
    \centering
    \small
    \begin{tabular}{p{0.1\textwidth}|p{0.35\textwidth}} \hline
    \textbf{Context}  &  \textbf{Sys} I have [value\_count] trains matching your request . Is there a specific day and time you would like to travel? \textbf{Usr} I would like to leave on [value\_day] and arrive by [value\_time]. \\  \hline
    \textbf{Model}           & \textbf{Generated Response} \\ \hline
    word RL:SL=off  & [train\_id] is leaving [value\_place] on [value\_day] on [value\_day] on [train\_id] [train\_id] [value\_count] [train\_id] leaving ...\\ \hline
    word RL:SL=100    & [train\_id] leaves at [value\_time] . would you like me to book you a ticket ? \\ \hline
    LiteAttnCat    & [train\_id] leaves [value\_place] at [value\_time] and arrives in [value\_place] at [value\_time]. Would you like me to book that for you ? \\ \hline
    \end{tabular}
    \caption{Example responses from baselines and LiteCatAttn on MultiWoz.} 
    \label{tbl:multiwoz_example}
\end{table}

\section{Model Analysis}
\label{sec:analysis}
\begin{figure}[ht]
    \centering
    \includegraphics[width=0.3\textwidth]{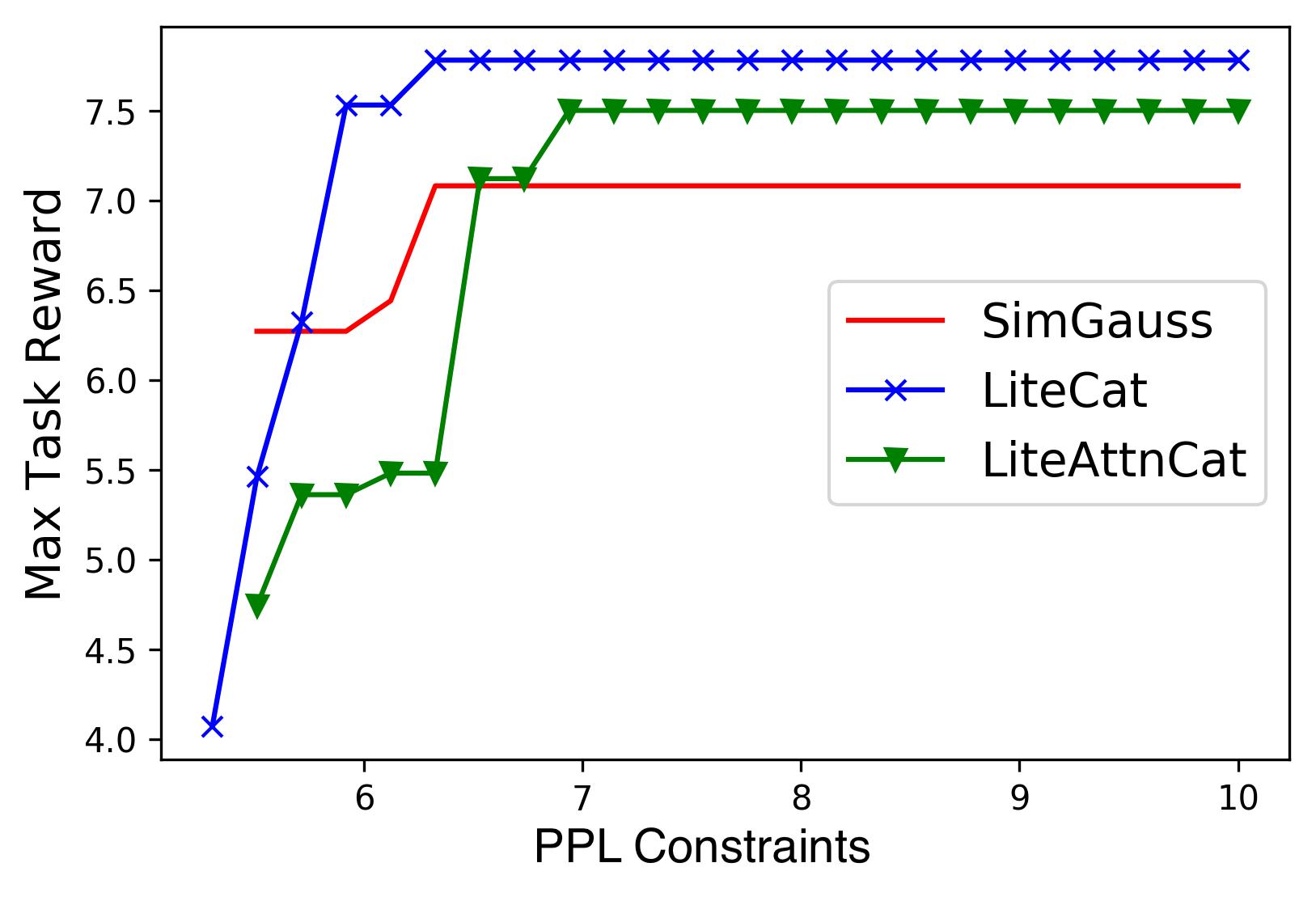}
    \includegraphics[width=0.3\textwidth]{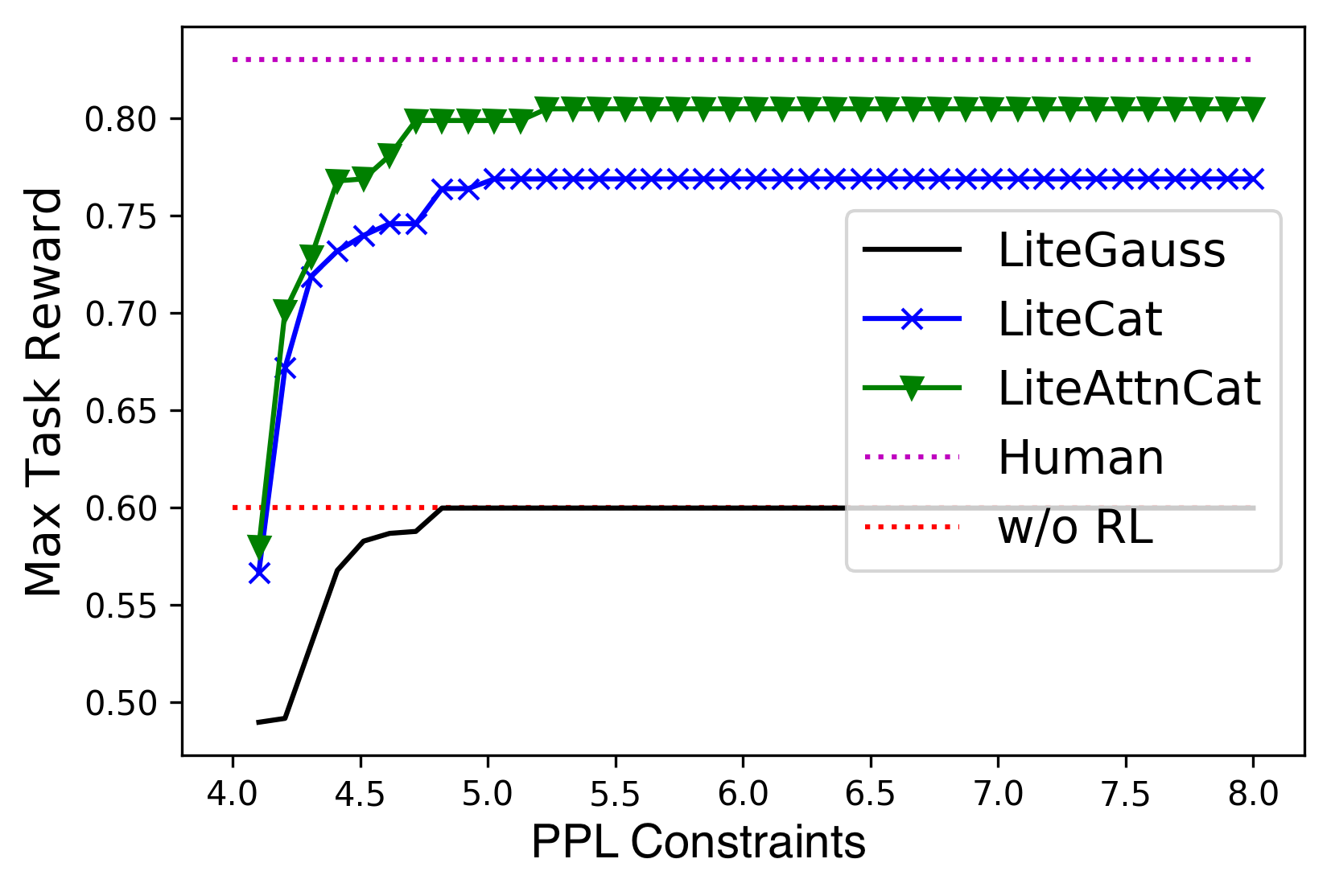}
    \caption{LCR curves on DealOrNoDeal and MultiWoz. Models with $\mathcal{L}_{full}$ are not included because their PPLs are too poor to compare to the Lite models.}
    \label{fig:rl_analysis}
\end{figure}

We compare the 6 variants of latent action models on DealOrNoDeal and MultiWoz. 
\begin{table}[ht]
\begin{tabular}{p{0.11\textwidth}|p{0.05\textwidth}p{0.06\textwidth}p{0.07\textwidth}p{0.07\textwidth}} \hline
\textbf{Deal}                & PPL    & Reward  & Agree\%   &  Diversity \\  \hline
Baseline        & 3.23   & 3.75   & 59      & 109    \\ \hline
Gauss           & 110K   & 2.71   & 43      & 176    \\ 
LiteGauss       & 5.35    & 4.48   & 65     & 91    \\   
Cat             & 80.41       & 3.9    & 62        & 115     \\ 
AttnCat         & 118.3        & 3.23    & 51         & 145        \\
LiteCat         & 5.35   & 2.67   & 41        & 58    \\ 
LiteAttnCat     & 5.25   & 3.69      & 52            &  75       \\\hline
\textbf{MultiWoz} & PPL    & BLEU  & Inform\%   &  Succ\% \\  \hline
Baseline        & 3.98   & 18.9   & 71.33      & 60.96    \\ \hline
Gauss             & 712.3   & 7.54   & 60.5      &  23.0   \\ 
LiteGauss         & 4.06     & 19.3   & 56.46   & 48.06      \\
Cat               & 7.07    & 13.7    & 54.15       &  42.04     \\ 
AttnCat           & 12.01    & 12.6    & 63.9        & 45.8     \\
LiteCat           & 4.10     & 19.1   & 61.56          & 49.15    \\ 
LiteAttnCat       & 4.05     & 19.1   & 67.97  & 57.36    \\ 
\hline 
\end{tabular}
\caption{Comparison of 6 model variants with only supervised learning training.}
\label{tbl:sl_analysis}
\end{table}
Table~\ref{tbl:sl_analysis} shows performance of the models that are pre-trained only with supervised learning. Figure~\ref{fig:rl_analysis} shows LCR curves for the 3 models pre-trained with $\mathcal{L}_{lite}$ and fine-tuned with policy gradient reinforcement learning.  
The following are the main findings based on these results.

\textbf{$\mathcal{L}_{lite}$ outperforms $\mathcal{L}_{full}$ as a pre-train objective.} Table~\ref{tbl:sl_analysis} shows that models with $\mathcal{L}_{full}$ fall behind their Lite counterparts on PPL and BLEU. We attribute this to the exposure bias in the latent space, i.e. the decoder is not trained to consider the discrepancy between the posterior network and actual dialog policy network. Meanwhile, the full models tend to enjoy higher diversity at pre-training, which agrees with the diversity-promoting effect observed in prior research~\cite{zhao2017learning}. However, our previous discussion on Figure~\ref{fig:deal_diversity} shows that Lite models are able to increase their response diversity in order to win more in negotiation through RL training. This is fundamentally different from diversity in pre-training, since diversity in LaRL is optimized to improve task reward, rather than to better model the original data distribution.
\begin{table}[ht]
    \centering
    \begin{tabular}{ccc|ccc} \hline
    $\beta$      & 0.0    & 0.01    & $\beta$      & 0.0    & 0.01  \\  \hline
    LiteCat       & 4.23   & 7.27   & LiteGauss       & 4.83   & 6.67  \\ \hline
    \end{tabular}
    \caption{Best rewards in test environments on DealOrNoDeal with various $\beta$.}
    \label{tbl:diff_beta}
\end{table}
Table~\ref{tbl:diff_beta} shows the importance of latent space regularization. When $\beta$ is $0$, both LiteCat and LiteGauss reach suboptimal policies with final reward that are much smaller than the regularized versions ($\beta=0.01$). The reason behind this is that the unregularized pre-trained policy has very low entropy, which prohibits sufficient exploration in the RL stage. 

\textbf{Categorical latent actions outperform Gaussian latent actions.} Models with discrete actions consistently outperform models with Gaussian ones. This is surprising since continuously distributed representations are a key reason for the success of deep learning in natural language processing. Our finding suggests that (1) multivariate categorical distributions are powerful enough to model complex natural dialog responses semantics, and can achieve on par results with Gaussian or non-stochastic continuous representations. (2) categorical variables are a better choice to serve as action spaces for reinforcement learning. Figure~\ref{fig:rl_analysis} shows that Lite(Attn)Cat easily achieves strong rewards while LiteGauss struggles to improve its reward. Also, applying REINFORCE on Gaussian latent actions is unstable and often leads to model divergence. We suspect the reason for this is the unbounded nature of continuous latent space: RL exploration in the continuous space may lead to areas in the manifold that are not covered in supervised training, which causes undefined decoder behavior given $z$ in these unknown areas.

\section{Conclusion and Future Work}
In conclusion, this paper proposes a latent variable action space for RL in E2E dialog agents. We present a general framework with a regularized ELBO objective and attention fusion for discrete variables. The methods are assessed on two dialog tasks and analyzed using the proposed LCR curve. Results show our models achieve superior performance and create a new state-of-the-art success rate on MultiWoz. Extensive analyses enable us to gain insight on how to properly train latent variables that can serve as the action spaces for dialog agents. This work is situated in the approach concerning practical latent variables in dialog agents, being able to create action abstraction in an unsupervised manner. We believe that our findings are a basic first step in this promising research direction.

\bibliography{naaclhlt2018}

\begin{thebibliography}{45}
\expandafter\ifx\csname natexlab\endcsname\relax\def\natexlab#1{#1}\fi

\bibitem[{Budzianowski et~al.(2018)Budzianowski, Wen, Tseng, Casanueva, Ultes,
  Ramadan, and Gasic}]{budzianowski2018multiwoz}
Pawe{\l} Budzianowski, Tsung-Hsien Wen, Bo-Hsiang Tseng, I{\~n}igo Casanueva,
  Stefan Ultes, Osman Ramadan, and Milica Gasic. 2018.
\newblock Multiwoz-a large-scale multi-domain wizard-of-oz dataset for
  task-oriented dialogue modelling.
\newblock In \emph{Proceedings of the 2018 Conference on Empirical Methods in
  Natural Language Processing}, pages 5016--5026.

\bibitem[{Cao and Clark(2017)}]{cao2017latent}
Kris Cao and Stephen Clark. 2017.
\newblock Latent variable dialogue models and their diversity.
\newblock In \emph{Proceedings of the 15th Conference of the European Chapter
  of the Association for Computational Linguistics: Volume 2, Short Papers},
  volume~2, pages 182--187.

\bibitem[{Chen et~al.(2013)Chen, Wang, and Rudnicky}]{chen2013unsupervised}
Yun-Nung Chen, William~Yang Wang, and Alexander~I Rudnicky. 2013.
\newblock Unsupervised induction and filling of semantic slots for spoken
  dialogue systems using frame-semantic parsing.
\newblock In \emph{Automatic Speech Recognition and Understanding (ASRU), 2013
  IEEE Workshop on}, pages 120--125. IEEE.

\bibitem[{Cho et~al.(2014)Cho, Van~Merri{\"e}nboer, Gulcehre, Bahdanau,
  Bougares, Schwenk, and Bengio}]{cho2014learning}
Kyunghyun Cho, Bart Van~Merri{\"e}nboer, Caglar Gulcehre, Dzmitry Bahdanau,
  Fethi Bougares, Holger Schwenk, and Yoshua Bengio. 2014.
\newblock Learning phrase representations using rnn encoder-decoder for
  statistical machine translation.
\newblock \emph{arXiv preprint arXiv:1406.1078}.

\bibitem[{Das et~al.(2017)Das, Kottur, Moura, Lee, and Batra}]{das2017learning}
Abhishek Das, Satwik Kottur, Jos{\'e}~MF Moura, Stefan Lee, and Dhruv Batra.
  2017.
\newblock Learning cooperative visual dialog agents with deep reinforcement
  learning.
\newblock In \emph{Computer Vision (ICCV), 2017 IEEE International Conference
  on}, pages 2970--2979. IEEE.

\bibitem[{Dhingra et~al.(2017)Dhingra, Li, Li, Gao, Chen, Ahmed, and
  Deng}]{dhingra2017towards}
Bhuwan Dhingra, Lihong Li, Xiujun Li, Jianfeng Gao, Yun-Nung Chen, Faisal
  Ahmed, and Li~Deng. 2017.
\newblock Towards end-to-end reinforcement learning of dialogue agents for
  information access.
\newblock In \emph{Proceedings of the 55th Annual Meeting of the Association
  for Computational Linguistics (Volume 1: Long Papers)}, volume~1, pages
  484--495.

\bibitem[{Gasic and Young(2014)}]{gasic2014gaussian}
Milica Gasic and Steve Young. 2014.
\newblock Gaussian processes for pomdp-based dialogue manager optimization.
\newblock \emph{IEEE/ACM Transactions on Audio, Speech, and Language
  Processing}, 22(1):28--40.

\bibitem[{Greensmith et~al.(2004)Greensmith, Bartlett, and
  Baxter}]{greensmith2004variance}
Evan Greensmith, Peter~L Bartlett, and Jonathan Baxter. 2004.
\newblock Variance reduction techniques for gradient estimates in reinforcement
  learning.
\newblock \emph{Journal of Machine Learning Research}, 5(Nov):1471--1530.

\bibitem[{He et~al.(2018)He, Chen, Balakrishnan, and Liang}]{he2018decoupling}
He~He, Derek Chen, Anusha Balakrishnan, and Percy Liang. 2018.
\newblock Decoupling strategy and generation in negotiation dialogues.
\newblock In \emph{Proceedings of the 2018 Conference on Empirical Methods in
  Natural Language Processing}, pages 2333--2343.

\bibitem[{Henderson et~al.(2014)Henderson, Thomson, and
  Young}]{henderson2014word}
Matthew Henderson, Blaise Thomson, and Steve Young. 2014.
\newblock Word-based dialog state tracking with recurrent neural networks.
\newblock In \emph{Proceedings of the 15th Annual Meeting of the Special
  Interest Group on Discourse and Dialogue (SIGDIAL)}, pages 292--299.

\bibitem[{Jang et~al.(2016)Jang, Gu, and Poole}]{jang2016categorical}
Eric Jang, Shixiang Gu, and Ben Poole. 2016.
\newblock Categorical reparameterization with gumbel-softmax.
\newblock \emph{arXiv preprint arXiv:1611.01144}.

\bibitem[{Kaelbling et~al.(1996)Kaelbling, Littman, and
  Moore}]{kaelbling1996reinforcement}
Leslie~Pack Kaelbling, Michael~L Littman, and Andrew~W Moore. 1996.
\newblock Reinforcement learning: A survey.
\newblock \emph{Journal of artificial intelligence research}, 4:237--285.

\bibitem[{Kingma and Welling(2013)}]{kingma2013auto}
Diederik~P Kingma and Max Welling. 2013.
\newblock Auto-encoding variational bayes.
\newblock \emph{arXiv preprint arXiv:1312.6114}.

\bibitem[{Kottur et~al.(2017)Kottur, Moura, Lee, and Batra}]{kottur2017natural}
Satwik Kottur, Jos{\'e} Moura, Stefan Lee, and Dhruv Batra. 2017.
\newblock Natural language does not emerge ‘naturally’in multi-agent
  dialog.
\newblock In \emph{Proceedings of the 2017 Conference on Empirical Methods in
  Natural Language Processing}, pages 2962--2967.

\bibitem[{Langford and Zhang(2008)}]{langford2008epoch}
John Langford and Tong Zhang. 2008.
\newblock The epoch-greedy algorithm for multi-armed bandits with side
  information.
\newblock In \emph{Advances in neural information processing systems}, pages
  817--824.

\bibitem[{Lee(2013)}]{lee2013structured}
Sungjin Lee. 2013.
\newblock Structured discriminative model for dialog state tracking.
\newblock In \emph{Proceedings of the SIGDIAL 2013 Conference}, pages 442--451.

\bibitem[{Lewis et~al.(2017)Lewis, Yarats, Dauphin, Parikh, and
  Batra}]{lewis2017deal}
Mike Lewis, Denis Yarats, Yann Dauphin, Devi Parikh, and Dhruv Batra. 2017.
\newblock Deal or no deal? end-to-end learning of negotiation dialogues.
\newblock In \emph{Proceedings of the 2017 Conference on Empirical Methods in
  Natural Language Processing}, pages 2443--2453.

\bibitem[{Li et~al.(2016)Li, Monroe, Ritter, Jurafsky, Galley, and
  Gao}]{li2016deep}
Jiwei Li, Will Monroe, Alan Ritter, Dan Jurafsky, Michel Galley, and Jianfeng
  Gao. 2016.
\newblock Deep reinforcement learning for dialogue generation.
\newblock In \emph{Proceedings of the 2016 Conference on Empirical Methods in
  Natural Language Processing}, pages 1192--1202.

\bibitem[{Liu and Lane(2017)}]{liu2017end}
Bing Liu and Ian Lane. 2017.
\newblock An end-to-end trainable neural network model with belief tracking for
  task-oriented dialog.
\newblock \emph{Proc. Interspeech 2017}, pages 2506--2510.

\bibitem[{Luong et~al.(2015)Luong, Pham, and Manning}]{luong2015effective}
Thang Luong, Hieu Pham, and Christopher~D Manning. 2015.
\newblock Effective approaches to attention-based neural machine translation.
\newblock In \emph{Proceedings of the 2015 Conference on Empirical Methods in
  Natural Language Processing}, pages 1412--1421.

\bibitem[{Mordatch and Abbeel(2017)}]{mordatch2017emergence}
Igor Mordatch and Pieter Abbeel. 2017.
\newblock Emergence of grounded compositional language in multi-agent
  populations.
\newblock \emph{arXiv preprint arXiv:1703.04908}.

\bibitem[{Raux et~al.(2005)Raux, Langner, Bohus, Black, and
  Eskenazi}]{raux2005let}
Antoine Raux, Brian Langner, Dan Bohus, Alan~W Black, and Maxine Eskenazi.
  2005.
\newblock Let’s go public! taking a spoken dialog system to the real world.
\newblock In \emph{in Proc. of Interspeech 2005}. Citeseer.

\bibitem[{Ren et~al.(2018)Ren, Xie, Chen, and Yu}]{ren2018towards}
Liliang Ren, Kaige Xie, Lu~Chen, and Kai Yu. 2018.
\newblock Towards universal dialogue state tracking.
\newblock In \emph{Proceedings of the 2018 Conference on Empirical Methods in
  Natural Language Processing}, pages 2780--2786.

\bibitem[{Serban et~al.(2017{\natexlab{a}})Serban, Sankar, Germain, Zhang, Lin,
  Subramanian, Kim, Pieper, Chandar, Ke et~al.}]{serban2017deep}
Iulian~V Serban, Chinnadhurai Sankar, Mathieu Germain, Saizheng Zhang, Zhouhan
  Lin, Sandeep Subramanian, Taesup Kim, Michael Pieper, Sarath Chandar,
  Nan~Rosemary Ke, et~al. 2017{\natexlab{a}}.
\newblock A deep reinforcement learning chatbot.
\newblock \emph{arXiv preprint arXiv:1709.02349}.

\bibitem[{Serban et~al.(2016)Serban, Sordoni, Bengio, Courville, and
  Pineau}]{serban2016building}
Iulian~V Serban, Alessandro Sordoni, Yoshua Bengio, Aaron Courville, and Joelle
  Pineau. 2016.
\newblock Building end-to-end dialogue systems using generative hierarchical
  neural network models.
\newblock In \emph{Proceedings of the 30th AAAI Conference on Artificial
  Intelligence (AAAI-16)}.

\bibitem[{Serban et~al.(2017{\natexlab{b}})Serban, Sordoni, Lowe, Charlin,
  Pineau, Courville, and Bengio}]{serban2017hierarchical}
Iulian~Vlad Serban, Alessandro Sordoni, Ryan Lowe, Laurent Charlin, Joelle
  Pineau, Aaron Courville, and Yoshua Bengio. 2017{\natexlab{b}}.
\newblock A hierarchical latent variable encoder-decoder model for generating
  dialogues.
\newblock In \emph{Thirty-First AAAI Conference on Artificial Intelligence}.

\bibitem[{Sordoni et~al.(2015)Sordoni, Galley, Auli, Brockett, Ji, Mitchell,
  Nie, Gao, and Dolan}]{sordoni2015neural}
Alessandro Sordoni, Michel Galley, Michael Auli, Chris Brockett, Yangfeng Ji,
  Margaret Mitchell, Jian-Yun Nie, Jianfeng Gao, and Bill Dolan. 2015.
\newblock A neural network approach to context-sensitive generation of
  conversational responses.
\newblock In \emph{Proceedings of the 2015 Conference of the North American
  Chapter of the Association for Computational Linguistics: Human Language
  Technologies}, pages 196--205.

\bibitem[{Su et~al.(2017)Su, Budzianowski, Ultes, Gasic, and
  Young}]{su2017sample}
Pei-Hao Su, Pawe{\l} Budzianowski, Stefan Ultes, Milica Gasic, and Steve Young.
  2017.
\newblock Sample-efficient actor-critic reinforcement learning with supervised
  data for dialogue management.
\newblock In \emph{Proceedings of the 18th Annual SIGdial Meeting on Discourse
  and Dialogue}, pages 147--157.

\bibitem[{Vinyals and Le(2015)}]{vinyals2015neural}
Oriol Vinyals and Quoc Le. 2015.
\newblock A neural conversational model.
\newblock \emph{arXiv preprint arXiv:1506.05869}.

\bibitem[{Walker(2000)}]{walker2000application}
Marilyn~A. Walker. 2000.
\newblock An application of reinforcement learning to dialogue strategy
  selection in a spoken dialogue system for email.
\newblock \emph{Journal of Artificial Intelligence Research}, pages 387--416.

\bibitem[{Wen et~al.(2016)Wen, Gasic, Mrksic, Rojas-Barahona, Su, Ultes,
  Vandyke, and Young}]{wen2016network}
Tsung-Hsien Wen, Milica Gasic, Nikola Mrksic, Lina~M Rojas-Barahona, Pei-Hao
  Su, Stefan Ultes, David Vandyke, and Steve Young. 2016.
\newblock A network-based end-to-end trainable task-oriented dialogue system.
\newblock \emph{arXiv preprint arXiv:1604.04562}.

\bibitem[{Wen et~al.(2017)Wen, Miao, Blunsom, and Young}]{wen2017latent}
Tsung-Hsien Wen, Yishu Miao, Phil Blunsom, and Steve Young. 2017.
\newblock Latent intention dialogue models.
\newblock In \emph{International Conference on Machine Learning}, pages
  3732--3741.

\bibitem[{Williams et~al.(2017)Williams, Asadi, and Zweig}]{williams2017hybrid}
Jason~D Williams, Kavosh Asadi, and Geoffrey Zweig. 2017.
\newblock Hybrid code networks: practical and efficient end-to-end dialog
  control with supervised and reinforcement learning.
\newblock In \emph{Proceedings of the 55th Annual Meeting of the Association
  for Computational Linguistics (Volume 1: Long Papers)}, volume~1, pages
  665--677.

\bibitem[{Williams and Young(2007)}]{williams2007partially}
Jason~D Williams and Steve Young. 2007.
\newblock Partially observable markov decision processes for spoken dialog
  systems.
\newblock \emph{Computer Speech \& Language}, 21(2):393--422.

\bibitem[{Williams and Zweig(2016)}]{williams2016end}
Jason~D Williams and Geoffrey Zweig. 2016.
\newblock End-to-end lstm-based dialog control optimized with supervised and
  reinforcement learning.
\newblock \emph{arXiv preprint arXiv:1606.01269}.

\bibitem[{Williams(1992)}]{williams1992simple}
Ronald~J Williams. 1992.
\newblock Simple statistical gradient-following algorithms for connectionist
  reinforcement learning.
\newblock \emph{Machine learning}, 8(3-4):229--256.

\bibitem[{Yang et~al.(2016)Yang, Yang, Dyer, He, Smola, and
  Hovy}]{yang2016hierarchical}
Zichao Yang, Diyi Yang, Chris Dyer, Xiaodong He, Alex Smola, and Eduard Hovy.
  2016.
\newblock Hierarchical attention networks for document classification.
\newblock In \emph{Proceedings of the 2016 Conference of the North American
  Chapter of the Association for Computational Linguistics: Human Language
  Technologies}, pages 1480--1489.

\bibitem[{Yarats and Lewis(2017)}]{yarats2017hierarchical}
Denis Yarats and Mike Lewis. 2017.
\newblock Hierarchical text generation and planning for strategic dialogue.
\newblock \emph{arXiv preprint arXiv:1712.05846}.

\bibitem[{Young et~al.(2007)Young, Schatzmann, Weilhammer, and
  Ye}]{young2007hidden}
Stephanie Young, Jost Schatzmann, Karl Weilhammer, and Hui Ye. 2007.
\newblock The hidden information state approach to dialog management.
\newblock In \emph{Acoustics, Speech and Signal Processing, 2007. ICASSP 2007.
  IEEE International Conference on}, volume~4, pages IV--149. IEEE.

\bibitem[{Young et~al.(2013)Young, Ga{\v{s}}i{\'c}, Thomson, and
  Williams}]{young2013pomdp}
Steve Young, Milica Ga{\v{s}}i{\'c}, Blaise Thomson, and Jason~D Williams.
  2013.
\newblock Pomdp-based statistical spoken dialog systems: A review.
\newblock \emph{Proceedings of the IEEE}, 101(5):1160--1179.

\bibitem[{Young(2006)}]{young2006using}
Steve~J Young. 2006.
\newblock Using pomdps for dialog management.
\newblock In \emph{SLT}, pages 8--13.

\bibitem[{Zhao and Eskenazi(2016)}]{zhao2016towards}
Tiancheng Zhao and Maxine Eskenazi. 2016.
\newblock Towards end-to-end learning for dialog state tracking and management
  using deep reinforcement learning.
\newblock In \emph{17th Annual Meeting of the Special Interest Group on
  Discourse and Dialogue}, page~1.

\bibitem[{Zhao and Eskenazi(2018)}]{zhao2018zero}
Tiancheng Zhao and Maxine Eskenazi. 2018.
\newblock Zero-shot dialog generation with cross-domain latent actions.
\newblock In \emph{Proceedings of the 19th Annual SIGdial Meeting on Discourse
  and Dialogue}, pages 1--10.

\bibitem[{Zhao et~al.(2018)Zhao, Lee, and Eskenazi}]{zhao2018unsupervised}
Tiancheng Zhao, Kyusong Lee, and Maxine Eskenazi. 2018.
\newblock Unsupervised discrete sentence representation learning for
  interpretable neural dialog generation.
\newblock In \emph{Proceedings of the 56th Annual Meeting of the Association
  for Computational Linguistics (Volume 1: Long Papers)}, volume~1.

\bibitem[{Zhao et~al.(2017)Zhao, Zhao, and Eskenazi}]{zhao2017learning}
Tiancheng Zhao, Ran Zhao, and Maxine Eskenazi. 2017.
\newblock Learning discourse-level diversity for neural dialog models using
  conditional variational autoencoders.
\newblock In \emph{Proceedings of the 55th Annual Meeting of the Association
  for Computational Linguistics (Volume 1: Long Papers)}, volume~1, pages
  654--664.

\end{thebibliography}
\bibliographystyle{acl_natbib}

\newpage
\appendix
\section{Supplemental Material}
\label{sec:supplemental}
\subsection{Training Details}
The following hyperparameters are used for the results on DealOrNoDeal.
\begin{table}[ht]
    \centering
    \small
    \begin{tabular}{p{0.15\textwidth}|p{0.25\textwidth}} \hline
    \multicolumn{2}{l}{\textbf{Supervised Pre-train}} \\ \hline
    Word Embedding      & 256 \\
    Utterance Encoder   & Attn GRU (128)  \\
    Context Encoder   & GRU (256)   \\
    Decoder        & GRU (256)   \\
    Optimizer      & Adam (lr=1e-3)  \\
    Dropout        & 0.5   \\ 
    $\beta$        & 0.01 \\ 
    Categorical $z$   & M=10, K=20\\ 
    Gaussian $z$       & M=200 \\ \hline
    \multicolumn{2}{l}{\textbf{Reinforce}} \\ \hline
    Optimizer  & SGD (lr=0.2 grad\_clip=0.1) \\
    $\gamma$     & 0.95 \\ \hline
    \end{tabular}
    \caption{Training details for DealOrNoDeal experiments. Attn GRU refers to~\cite{yang2016hierarchical}}
\end{table}

The following hyperparameters are used for the results on MultiWoz.
\begin{table}[ht]
    \centering
    \small
    \begin{tabular}{p{0.15\textwidth}|p{0.25\textwidth}} \hline
    \multicolumn{2}{l}{\textbf{Supervised Pre-train}} \\ \hline
    Word Embedding      & 256 \\
    Encoder   & Attn GRU (300)  \\
    Decoder        & LSTM (150)   \\
    Optimizer      & Adam (1e-3)  \\
    Dropout        & 0.5   \\ 
    $\beta$        & 0.01 \\ 
    Categorical $z$ & M=10, K=20\\ 
    Gaussian $z$ & M=200 \\ \hline
    \multicolumn{2}{l}{\textbf{Reinforce}} \\ \hline
    Optimizer  & SGD(lr=0.01 grad\_clip=0.5) \\
    $\gamma$     & 0.99 \\ \hline
    \end{tabular}
    \caption{Training details for MultiWoz experiments}
\end{table}

\subsection{DealOrNoDeal Example Conversations}
\label{appendix:example}
The followings are examples dialogs generated from word-level models (Table~\ref{tbl:deal_example}) and latent-level models (Table~\ref{tbl:deal_latent_example}).
\begin{table}[ht]
    \centering
    \small
    \begin{tabular}{p{0.45\textwidth}} \hline
    \textbf{Scenario 1} \\ \hline
    1 book, 1 hat and 3 balls available. \\
    \textbf{Value Function} \\ 
    Agent: book=1, hat=6, ball=1. \\
    User: book=1, hat=6, ball=1. \\
    \textbf{Dialog} \\
    Agent: I'd like the hat and 2 balls. \\
    User: Deal. \\
    Agent: SELECTION \\
    \textbf{Outcome: Agreement} \\
    Agent: 1 hat and 2 balls; User: 1 book and 1 ball \\
    Agent reward = 8; User reward = 2 \\ \hline
    \textbf{Scenario 2} \\ \hline
    4 books, 1 hat and 1 ball available. \\
    \textbf{Value Function} \\ 
    Agent: book=0, hat=7, ball=3; \\
    User: book=1, hat=6, ball=0.\\
    \textbf{Dialog} \\
    User: I can offer you 1 hat and 2 books. \\
    Agent: I need the hat and ball.\\
    User: That won't work for me. I can offer you one book and the hat.\\
    Agent: I need the hat and ball.\\
    User: I have to have the hat.\\
    Agent: I need the hat and ball. \\
    User: Okay. \\
    Agent: SELECTION \\
    \textbf{Outcome: Agreement} \\
    Agent: 1 hat and 1 ball; User: 4 books \\
    Agent reward = 10; User reward = 4 \\ \hline
    \end{tabular}
    \caption{Example dialogs between baseline with the user model. Agent is trained with word-level policy gradient and the user is a supervised pre-trained model.}
    \label{tbl:deal_example}
\end{table}

\begin{table}[ht]
    \centering
    \small
    \begin{tabular}{p{0.45\textwidth}} \hline
    \textbf{Scenario 1} \\ \hline
    2 books, 2 hats and 1 ball available. \\
    \textbf{Value Function} \\ 
    Agent: book=3, hat=0, ball=4; \\
    User: book=2, hat=3, ball=0. \\
    \textbf{Dialog} \\
    Agent: Can i have the books and one ball? \\
    User: I would like the books and the hats then. \\
    Agent: I would like the ball and books. \\
    User: Deal. \\
    Agent: SELECTION\\
    \textbf{Outcome: Agreement} \\
    Agent: 2 books and 1 ball \\
    User: 2 hats \\
    Agent reward = 10, User reward = 6 \\ \hline
    \textbf{Scenario 2} \\ \hline
    2 books, 1 hat and 3 ball available. \\
    \textbf{Value Function} \\ 
    Agent: book=0, hat=4, ball=2; \\
    User: book=3, hat=1, ball=1.\\
    \textbf{Dialog} \\
    Agent: I would like the hat and balls.\\
    User: I cannot accept that. I need the hat and books.\\
    Agent: Can I have the hat and balls, you get the rest?\\
    User: Okay, deal.\\
    Agent: SELECTION. \\
    \textbf{Outcome: Agreement} \\
    Agent: 1 hat and 3 ball \\
    User: 2 books \\
    Agent reward = 10, Simulator reward = 6 \\ \hline
    \end{tabular}
    \caption{Example dialogs between LiteCat and the user model. Agent is trained with latent-level policy gradient and the user is a supervised pre-trained model.}
    \label{tbl:deal_latent_example}
\end{table}

\end{document}